# Learning from Sparse Datasets: Predicting Concrete's Strength by Machine Learning

**Boya OUYANG (1, 2), Yuhai LI (1), Yu SONG (1), Feishu WU (1), Huizi YU (1), Yongzhe WANG (1), Mathieu BAUCHY (1, 4) and Gaurav SANT (2, 3, 4)**

(1) Physics of AmoRphous and Inorganic Solids Laboratory (PARISlab), Department of Civil and Environmental Engineering, University of California, Los Angeles, CA 90095, USA

(2) Department of Materials Science and Engineering, University of California, Los Angeles, CA, USA

(3) Laboratory for the Chemistry of Construction Materials (LC$^2$), Department of Civil and Environmental Engineering, University of California, Los Angeles, CA, USA

(4) Institute for Carbon Management, University of California, Los Angeles, CA, USA

**Abstract**
Despite enormous efforts over the last decades to establish the relationship between concrete proportioning and strength, a robust knowledge-based model for accurate concrete strength predictions is still lacking. As an alternative to physical or chemical-based models, data-driven machine learning (ML) methods offer a new solution to this problem. Although this approach is promising for handling the complex, non-linear, non-additive relationship between concrete mixture proportions and strength, a major limitation of ML lies in the fact that large datasets are needed for model training. This is a concern as reliable, consistent strength data is rather limited, especially for realistic industrial concretes. Here, based on the analysis of a large dataset (>10,000 observations) of measured compressive strengths from industrially-produced concretes, we compare the ability of select ML algorithms to "learn" how to reliably predict concrete strength as a function of the size of the dataset. Based on these results, we discuss the competition between how accurate a given model can eventually be (when trained on a large dataset) and how much data is actually required to train this model.

## 1. INTRODUCTION

The 28-day compressive strength is one of the most widely accepted metrics to characterize concrete's performance for engineering applications. Indeed, although this standardized yet simple index is primarily used to evaluate the ultimate strength of concrete mixtures [1], it can also serve as an expedient measure to infer other critical mechanical properties such as elastic modulus, stiffness, or tensile strength [2]. Accurate strength predictions in concrete design have a profound impact on the efficiency and quality of construction projects. Indeed, for instance, an insufficient concrete strength can be the culprit of a catastrophic failure of civil infrastructures. Conversely, concretes exhibiting an overdesigned strength leads not only to higher material expenses [3], but also to additional environmental burdens—such as $CO_2$ emissions in cement production [4].

Over the past decades, a substantial amount of effort has been devoted to developing predictive models for correlating a given concrete mixture proportion to its associated strength performance [5]. Beyond this, an ideal predictive model also provides important insights for designing new concrete with better constructability and durability, and/or at a lower cost [6,7]. Conventional approaches often seek to achieve these goals using physics or chemistry-based relationships [8–10]. Although the role played by major proportioning parameters (e.g., water-to-cementitious ratio, w/cm, aggregate fraction, and air void content) has been extensively investigated, the influence of many other factors is not always negligible, e.g., chemical and mineral admixtures or aggregates gradation [11]. Due to the limited understanding of these complex property-strength correlations, it is still extremely challenging to get a robust and universal concrete strength model using conventional approaches [12].

As an alternative pathway, the recent development of machine learning (ML) techniques provides a novel data-driven approach to revisit the strength prediction problem. Importantly, ML-based predictions have been shown to significantly outperform those of conventional approaches, especially when handling non-linear problems [13]. Without the need for any physical or chemical presumptions, this new approach also further permits greater flexibility to extract hidden, non-intuitive feature patterns directly from the input data. As such, recent studies have established ML as a promising approach to predict concrete strength[14–17]. However, a major limitation of ML approaches lies in the fact that a large dataset is usually required for ML algorithms to "learn" the relationship between inputs and outputs [18,19]. This is a major concern for concrete strength applications, as strength data for industrial concretes are often difficult to access (i.e., data is not publicly available). In addition, reported concrete strength data are often incomplete, that is, some important features are often missing, e.g., curing temperature, additives, types of aggregates, etc. More generally, ML approaches require accurate and self-consistent data—which is often questionable for concrete strength data due to non-standardized measurements or inconsistencies in data recording [20]. For example, the strength of a given concrete material can significantly vary when the testing protocol or specimen size is changed [21–23]. Although such difficulties can be filtered out with sufficiently large datasets, their significance tends to be exacerbated in the case of small datasets. For all these reasons, it is critical to assess how the reliability of ML approaches for concrete strength prediction applications depends on the number of training data points.

This study revolves around two core questions: (i) how much data is sufficient for training a ML model and (ii) which ML algorithms are better suited to deal with small datasets. Here, by building on our previous studies [17,24], we explore the above questions by taking the example of three archetypal learning algorithms, namely, polynomial regression (PR), artificial neural network (ANN), and random forest (RF). We compare the ultimate learning accuracy of these algorithms (i.e., based on the entire training set), as well as their learning efficiency as a function of data volume. These results are insightful for facilitating the adoption of ML techniques for small datasets—as relevant to concrete engineering.

## 2. BACKGROUND AND METHODS

### 2.1 Machine learning algorithms

We assess the performance of three common, archetypical learning algorithms (PR, ANN, and RF) as a function of the number of training data points. These methods are chosen as they belong to three distinct families of ML models, namely, polynomial, network-based, and tree-based [25,26]. Note that all the hyperparameters of the ML models considered herein were optimized in a previous study so as to achieve an optimal balance between under- and overfitting [16]. First, we consider PR, which is essentially based on linear regression, wherein the model parameters designate an *n*-degree polynomial function [27]. Based on our previous work [16], the PR model adopted herein features a maximum polynomial degree of 3. Second, we explore the potential ANN, which is a computational structure consisting of an input layer, an output layer, and one or several hidden layers bridging the two formers—wherein each layer comprises a collection of artificial neurons (i.e., computational units) [28]4/29/20 11:02:00 AM. Based on our previous work [16], the present ANN model exhibits 7 neurons in a single hidden layer. We adopt the sigmoid function as activation function to prioritize the importance of the input data and we use the backpropagation algorithm to optimize the model parameters [29]. Third, we consider RF, which is an enhanced bagging method since, by using the majority-voting concept, this approach is typically more predictive than conventional decision trees [30]. Here, based on our previous work [16], our RF model comprises 16 trees. Despite the different nature of these algorithms, their common goal is to predict a variable *y* (i.e., the 28-day strength) as a function of the input variables *x* (i.e., mixing proportions of concrete), while minimizing the difference between measured and predicted strength values (see Ref. [16] for details).

## 2.2 Feature selection

The dataset used in this study includes the 28-day compressive strength of 10,264 commercial concretes and associated mixture proportions [17]. All the mixtures were cast using ASTM C150 compliant Type I/II cement [31] and Class F fly ash compliant with ASTM C618 [32]. The seven most influential features are considered in this study, namely, (1) w/cm, (2) cement %, (3) fly ash %, (4) fine aggregate %, (5) air-entraining admixture (AEA) dosage, and (6) water-reducing admixture (WRA) dosage. For normalization purposes, the features from (2) to (4) are taken as the solid weight fractions, wherein the fraction of coarse aggregates is excluded as it is redundant (i.e., the sum of all the weight fractions is 100%).

## 2.3 Model training

Following common practices in ML, 70% of the strength observations are randomly selected and used for model training (i.e., "training set"). The remaining 30% of the data are kept hidden to the model and assess the ability of the model to predict the strength of unknown concretes (i.e., "test set"). The hyperparameters of each mode are optimized by five-fold cross-validation [33]. In detail, the training set is randomly split into five smaller folds (each made of 20% of the training data). In each of the five rounds of analysis, the model is iteratively trained based on four folds and validated based on the remaining fold (i.e., "cross-validation set").

## 2.4 Accuracy evaluation

We evaluate the accuracy of each model by calculating their mean-square error (MSE) and coefficient of determination ($R^2$), wherein the MSE is the averaged Euclidian distance between predicted and measured strength data in the test set. The relative MSE (RMSE) is then calculated as the square root of the MSE. The $R^2$ factor further quantifies the accuracy of the model predictions in terms of the degree of scattering around the fitted input-output relationship (a perfect prediction would be associated with $R^2 = 1$. We further analyze the deviation between strength predictions and measurements by computing the error distribution—that is, the distribution of the differences between predicted and measured strength values for each concrete mixture in the test set. The error distribution yielded by each model then serves to calculate the 90 and 95% confidence intervals of a predicted strength falling into these ranges (see Ref. [16] for details).

## 2.5 Evaluation of the learning efficiency

To investigate how each model "learns" how to predict concrete strength as it exposed to more training examples, we compute their "learning curve" [34]. This approach consists of plotting the accuracy of the model as it is exposed to an increasing number of training examples. Here, we compute the MSE (for the training and validation sets) while gradually increasing the size of the training set by 10% increments. To ensure consistent comparison, all the models are trained and evaluated based on identical training and validation sets.

## 3. RESULTS AND DISCUSSION

## 3.1 Accuracy of the machine learning models

We first compare the final accuracy offered by each ML model, that is, when trained based on the entire training set. To this end, Fig. 1 shows for each model the predicted vs. measured strengths for the entire test set, as well as the associated error distributions. The accuracy analysis is summarized in Tab. 1. In detail, we find that RF features the highest degree of accuracy, which manifests itself by a minimum RMSE, maximum $R^2$, and minimum confidence intervals.

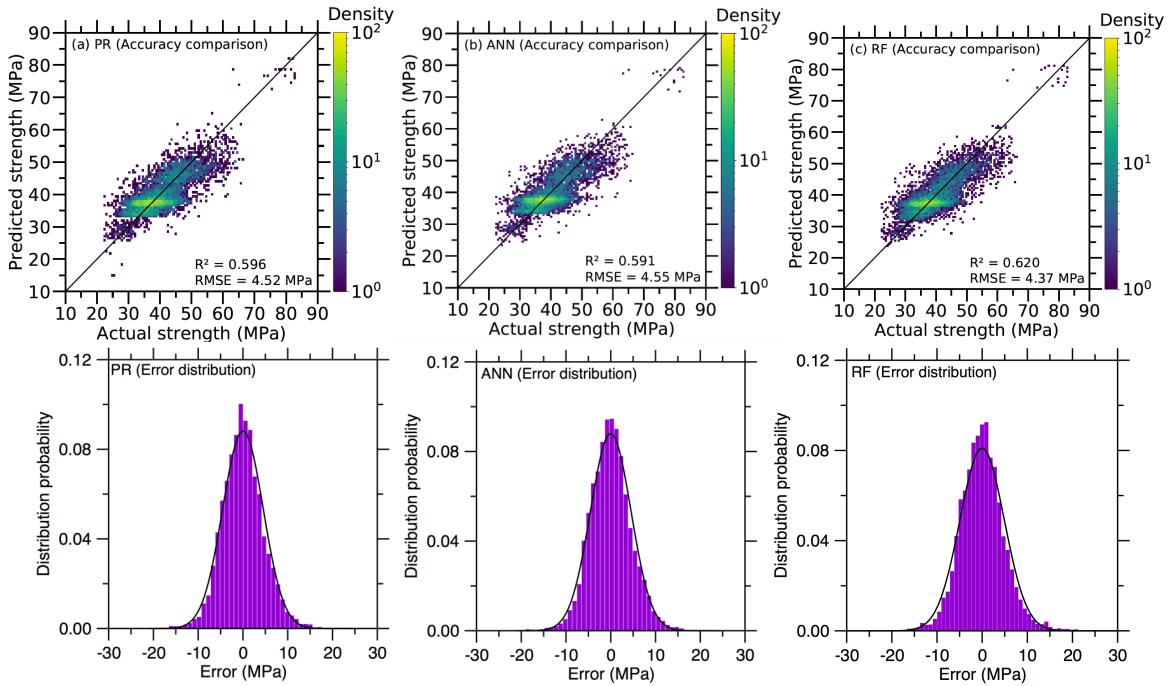

**Figure 1: Comparison between predicted vs. measured (ground-truth) strengths (top) and error distribution (bottom) for the (a) PR, (b) ANN, and (c) RF models. The pixel colors in the left plots indicate the number of overlapped points. The error distributions are fitted by a Gaussian distribution function.**

## 3.2 Gradual learning upon increasing training set size

Having shown that RF offers the best final accuracy when trained based on the entire training set, we now focus on the learning curve exhibited by each model—to assess their ability to quickly learn the input-output relationship as they become exposed to a gradually increasing number of training examples, as shown in Fig. 2. As expected, all the models exhibit a fairly similar trend, that is, (i) the MSE of the training set increases with increasing training set size since it becomes increasingly difficult from the model to perfectly interpolate the training set and (ii) the MSE of the cross-validation set decreases with increasing training set size as the model gradually manages to learn the input-output relationship and, hence, eventually shows an increased ability to predict the strength of unknown concretes.

**Table 1: Values of $R^2$ and confidence intervals over the test set for each model (when trained based on the entire training set) and minimum number of training data that is needed for each model to achieve an average validation set MSE that is less than one standard deviation away from its final validation set MSE.**

| Model type | Accuracy analysis | | | Learning analysis |
|---|---|---|---|---|
| | $R^2$ | Confidence interval (MPa) | | Minimum number of training data points to reach maximum accuracy |
| | | 90% | 95% | |
| PR | 0.596 | ± 7.43 | ± 8.86 | 2680 |
| ANN | 0.591 | ± 7.45 | ± 8.88 | 3010 |
| RF | 0.620 | ± 7.22 | ± 8.60 | 4070 |

Nevertheless, we find that, although the final accuracy offered by the models shows only minor differences (see Tab. 1), their learning curves exhibit more significantly distinct features. In detail, in agreement with the data presented in Tab. 1, we find that RF eventually features the lowest MSE for the validation set, as well as for the training set. However, we note that the MSE of the validation set exhibits a faster decrease in the case of PR and ANN. We further quantify this behavior by computing the minimum number of training data points that is needed for the model to achieve an average validation set MSE that is less than one standard deviation away from its final validation set MSE (i.e., when trained based on the entire training set), wherein the standard deviation is calculated based on the MSE obtained for each validation fold in cross-validation. Overall, we find that PR and, to a lesser extent, ANN features an increased ability to quickly learn how to predict concrete strength from small datasets as compared to RF (see Tab. 1).

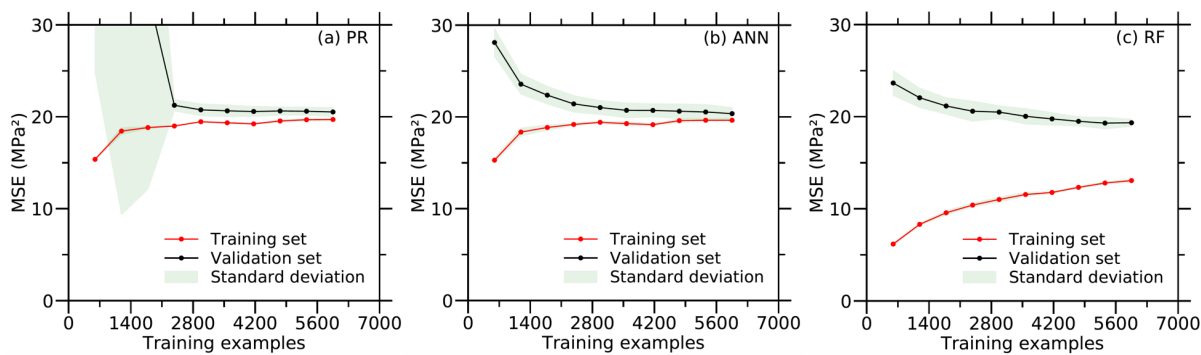

**Figure 2: Learning curves showing the MSE of the training and cross-validation sets as a function of the size of the training set for the (a) PR, (b) ANN, and (c) RF models.**

### 3.3 Competition between model accuracy and need for large dataset

Overall, we find that the model offering the highest final degree of accuracy (i.e., RF) requires the largest training set to be trained, whereas, in turn, the models presenting the lowest final accuracy (i.e., PR and ANN) require the smallest training set to be trained. These results suggest the existence of competition between (i) the final ability of a model to accurately learn the input-output relationship when trained based on an excess of training examples and (ii) the ability of a model to quickly learn this relationship when trained based on a small dataset. This competition can be rationalized in terms of the intrinsic "flexibility" of the model.

On the one hand, PR and ANN are constrained, poorly-flexible models—since PR relies on a fixed analytical form, while the present ANN model exhibits a limited ability to capture complex input-output relationships as it comprises a single hidden layer. This lack of flexibility limits the final accuracy that is achievable by these models. Although the degree of complexity of these models (i.e., maximum polynomial degree for PR and number of hidden neurons for ANN) is already tuned to achieve the best balance between under- and overfitting (see Ref. [16]), the fact that the MSE of the training and validation sets both plateau toward the same value suggests that these models are too simple and lack some degrees of freedom. For a given amount of data, this limitation could potentially be mitigated by carefully increasing the complexity of these models (while avoiding overfitting)—for instance, by increasing the number of hidden layers in ANN [35]. In turn, the constrained nature of these models allows them to quickly achieve their maximum accuracy—since only a limited number of parameters (i.e., polynomial coefficients for PR and neuron-neuron connection weights for ANN) need to be parameterized [36]. This makes it possible for these algorithms to handle small datasets. However, it is clear from Fig. 2 that these models have already achieved their maximum accuracy and, hence, would not benefit from being trained with any additional data.

On the other hand, RF is, in contrast, more flexible as it is not constrained by any analytical formulation. Indeed, in contrast to PR (which intrinsically yields a smooth, continuous, and differentiable relationship between inputs and output due to its analytical form), the tree-based structure makes it possible for the RF model to capture rough, less continuous/differentiable functions [37]. This flexibility enables RF to eventually reach a higher final degree accuracy once trained based on the entire training set. In turn, such complexity comes at a cost, namely, a large number of training data points is needed to properly parameterize the RF model. This is well illustrated by the facts that, unlike the cases of PR and ANN, (i) the validation set MSE of the RF model does not reach a plateau and continues to decrease upon increasing training set size and (ii) the final validation set MSE is significantly higher than the final training set MSE. Both of these learning curve features suggests that the RF model has not yet finished its training and, hence, could further by improved if exposed to an increased number of data—that is, unlike the PR and ANN models, the RF model still features some room for improvement

## 5. CONCLUSIONS

- Machine learning offers a promising pathway to predict concrete strength.
- Simple, more constrained models (e.g., PR) offer limited final accuracy, but can quickly achieve their maximum accuracy while trained based on a small training set.
- Less constrained, more flexible models (e.g., RF) require larger training sets, but can eventually feature a higher final prediction accuracy.

## ACKNOWLEDGEMENTS

The authors acknowledge some financial support for this research provided by the U.S. Department of Transportation through the Federal Highway Administration (Grant #: 693JJ31950021) and the U.S. National Science Foundation (DMREF: 1922167).